# A Multi-Strategy Framework for Enhancing Shatian Pomelo Detection in Real-World Orchards

Pan Wang[a+], Yihao Hu[a+], Xiaodong Bai[a*], Jingchu Yang[a], Leyi Zhou[a], Aiping Yang[b], Xiangxiang Li[b], Meiping Ding[b], Jianguo Yao[c]

[a] School of Computer Science and Technology, Hainan University, Haikou 570228, Hainan, China.

[b] Agricultural Meteorological Center, Jiangxi Meteorological Bureau, Nanchang 330045, Jiangxi, China.

[c] School of Communications and Information Engineering, Nanjing University of Posts and Telecommunications, Nanjing 210003, Jiangsu, China.

[+]These authors contributed to the work equally.

*Correspondence: xiaodongbai@hainanu.edu.cn (X. Bai); Tel.: +86 0898 6627 1330.

## Abstract

Shatian pomelo, as a specialty agricultural product with a large market, requires automated detection systems to ensure accurate quantity assessment and meet commercial demands for lean production. Existing detection methodologies often employ specialized networks tailored for idealized theoretical conditions or specific dataset characteristics, yet these approaches tend to exhibit significant performance degradation in real-world orchard environments. This study identifies four challenges impacting detection accuracy: imaging devices, lighting conditions, object scale variation, and occlusion. To mitigate these challenges, a multi-strategy framework is proposed. Firstly, a multi-scenario dataset, STP-AgriData, is developed by integrating real orchard images with internet-sourced data to address tone variation arising from different imaging devices and complex orchard conditions. Secondly, to simulate inconsistent illumination conditions, data augmentations such as contrast adjustment and brightness modification are applied. Thirdly, to more effectively handle scale variation and occlusion, we propose a novel detection network, REAS-Det. Specifically, we design Global-Selective Visibility Convolution (GSV-Conv). Unlike conventional convolutions, GSV-Conv enhances feature aggregation by selectively expanding the visible feature space under global semantic awareness. Meanwhile, it preserves efficient spatial aggregation, thereby enabling adaptive perception across varying object scales. In addition, we incorporate the C3RFEM module, the multi-scale multi-head feature selection mechanism (MultiSEAM), and Soft-NMS to enhance instance separation and localization accuracy. Experimental results achieved a Precision of 86.5%, a Recall of 77.2%, a mAP@.50 of 84.3%, and a mAP@.50:.95 of 53.6%. The proposed network outperforms other state-of-the-art detection methods, achieving high detection accuracy in real orchard environments. The source code will be available at: https://github.com/Genk641/REAS-Det.

## 1. Introduction

Shatian pomelo is noted for its delicious taste and high nutritional value, with an annual output of roughly 360,000 tons in Guangxi, China (Newswire, 2023). As agricultural e-commerce grows, a substantial portion of Shatian pomelos is now sold online (Yang, Liu, Cao, Sun, & Kong, 2024). Accurate fruit yield estimation is essential for online sales platforms, as it plays a vital role in optimizing logistics and determining warehouse storage strategies. Moreover, with the rapid development of the modern insurance industry, many agricultural insurance policies have been introduced to compensate orchards for losses caused by natural disasters. Consequently, to facilitate accurate insurance assessments, it is imperative to estimate fruit counts in damaged orchards rapidly and accurately. However, current methods for Shatian pomelo counting mainly rely on manual sampling and predictive estimation, which are inefficient and susceptible to subjective factors, often resulting in significant discrepancies from actual results (Kamilaris & Prenafeta-Boldú, 2018). In addition, rising labor costs in recent years have made manual counting increasingly expensive, driving up production costs. Hence, using automated counting methods to replace manual labor has become a new trend, which relies heavily on vision algorithms for accurate detection and counting (S. Liu, Zeng, & Whitty, 2020). However, achieving high precision, especially in complex orchard environments, remains a significant challenge. Therefore, how to improve the accuracy of fruit detection has drawn substantial research attention.

Current research has demonstrated the effectiveness of computer vision methods in detecting common fruits such as apples (M. Sun, Xu, Luo, Lu, & Jia, 2022), cherry tomatoes (W. Chen, Liu, Zhao, Li, & Wang, 2024), strawberries (Zhou, et al., 2024), and citrus fruits (Y. Zhang, et al., 2024). These methods can be broadly categorized into two groups: traditional machine learning methods and deep learning-based methods. Traditional machine learning methods are based on manually designed features. For example, Liu et al. (T.-H. Liu, Ehsani, Toudeshki, Zou, & Wang, 2018) proposed a computer vision algorithm to identify ripe pomelos using an elliptical boundary model. The image was first converted from RGB space to Y'CbCr space. Then, an Ordinary Least Squares (OLS) method was introduced in the Cr-Cb color space to fit an implicit second-order polynomial of the elliptical boundary models, which enabled the segmentation of various fruits. Yan et al. (Yan, et al., 2021) used contour coordinate transformation fitting to extract fruit shape features. By implementing an orientation angle compensation algorithm, they achieved rapid and accurate measurement of pomelo vertical and horizontal dimensions, enabling precise fruit size and shape determination. For citrus belonging to the same genus as pomelos, a morphology-based Multi-class Support Vector Machine (SVM) was used to segment fruit from branches under sunny and cloudy conditions, achieving an accuracy of 92.4% (Qiang, Jianrong, Bin, Lie, & Yajing, 2014). Xu et al. (Xu, et al., 2020) applied the Otsu adaptive threshold method for citrus segmentation and the Canny operator for edge extraction, which raised the overall recognition rate of citrus to 95%. From these examples, it is evident that traditional machine learning methods can indeed achieve fruit detection by recognizing simple features such as shape and color. However, they still fail to fully extract high-dimensional features of fruits. They are limited by accuracy and adaptability when dealing with complex orchard images. Moreover, these limitations are further intensified in orchards due to factors such as changes in lighting, similar backgrounds, foliage, and shadow coverage, which affect the surface features of fruits.

Deep learning-based computer vision represents one of the modern techniques for image processing, exhibiting strong robustness and generalization capabilities. Nowadays, it plays a key role in smart agriculture, such as crop pest and disease detection (J. Liu & Wang, 2021), intelligent irrigation (Zeng, Chen, & Lin, 2023), yield prediction (Moussaid, et al., 2023), and precision agriculture (Ganatra & Patel, 2021). Particularly for agricultural object detection, this method can effectively capture multi-scale contextual information, extract relevant features, and enhance generalization. Its effectiveness is evidenced by remarkable results across diverse applications (Oppenheim, Shani, Erlich, & Tsror, 2019; J. Qi, et al., 2022; J. Wang, et al., 2023). Current deep learning-based methods fall into two main categories: two-stage convolutional neural networks based on candidate regions and one-stage convolutional neural networks using regression. For citrus plants similar to Shatian pomelo, Wen et al. (Wen, et al., 2020) improved the Mask R-CNN framework by using two-stage detection methods, and achieved a multi-task detection mAP of 91.56% in complex environments. In comparison, one-stage detection methods exhibit lower accuracy but greater computational efficiency, making them widely suitable for real-time fruit detection tasks. For example, Juneja et al. proposed (Juneja, Juneja, Soneja, & Jain, 2021) a CNN-based Single Shot Multibox Detector (SSD) for real-time object detection in previous research, which achieved a real-time detection accuracy of 92.7%. Tian et al. (Tian, et al., 2019) used an improved YOLOv3 model to recognize apples at different growth stages, employing DenseNet to process low resolution feature layers, which enhanced the detection accuracy of apples in the orchard. Based on YOLOv4, Gai et al. (Gai, Chen, & Yuan, 2023) incorporated CSPDarknet53 and DenseNet layers, and employed circular marker boxes to optimize cherry fruit detection for precision agriculture. Similarly, using YOLO as a foundation, Lyu et al. (S. Lyu, et al., 2022) introduced an attention module and modified the loss function, which improved this model's accuracy in counting green citrus in complex orchards. For occluded green fruits, Jia et al. (Jia, et al., 2022) designed a fast optimized Foveabox Detection Model (Fast-FDM) for quick recognition and localization of green apples, achieving an average detection accuracy of 62.3%. Moreover, Liu et al. (M. Liu, et al., 2022) proposed a Fully Convolutional One-Stage (FCOS) object detection model, improved the feature pyramid network, and added a convolutional block attention network, achieving a detection accuracy of 81.2%. In real-world scenarios, edge devices like picking robots must meet stringent real-time requirements with limited computing resources. Thus, a one-stage convolutional neural network is more suitable for complex orchard environments.

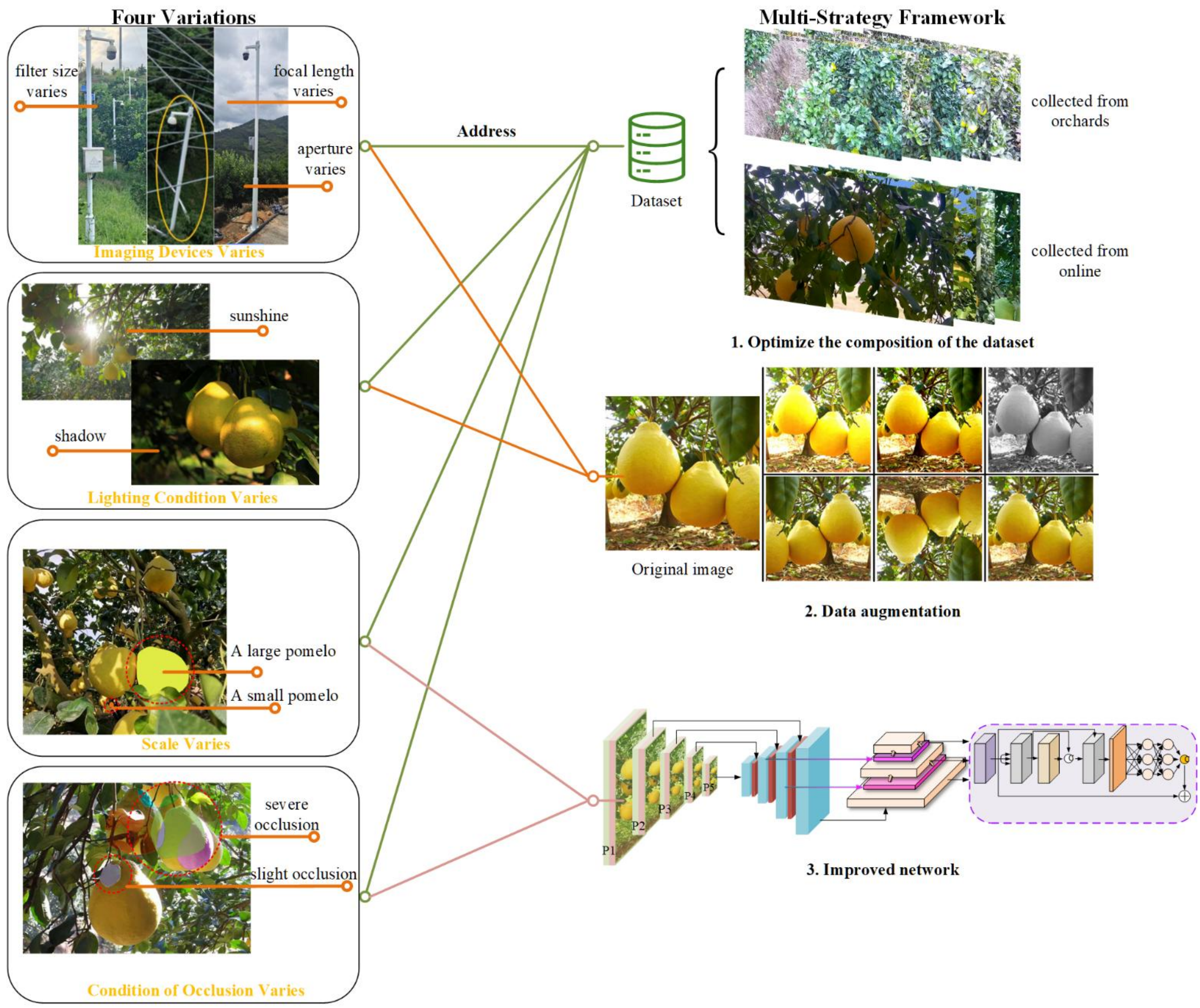

Fig. 1 A Comprehensive Multi-Strategy Framework. Four variations and a corresponding multi-strategy framework. The connections between variations and framework represent effective approaches.

Although the aforementioned studies have showcased many successful examples of using one-stage methods in fruit detection, they often encounter degraded performance in real-world applications. Specifically, the existing research typically introduces specialized network architectures tailored for distinct scenarios encompassing single imaging devices, specific orchard environments, and similar lighting conditions. However, in practical applications, imaging devices are varied, field environments are changeable, lighting conditions are unstable, and fruit postures are markedly different. These factors lead to poor application results of this scheme in reality. To address the requirements of real-world application scenarios, Shatian pomelos were analyzed as samples, revealing four key variations between practical and laboratory applications. First, the diversity of imaging devices often leads to significant variations in image quality. The variations in sharpness and color saturation caused by imaging devices can influence the effectiveness of fruit detection methods. Second, the lighting condition varies. Fruit surface features become unclear when exposed to light. Third, the scale varies. Shatian pomelo trees are densely planted in orchards. During image capture, fruit at varying distances (from far to near) may appear in a single image, which results in significant scale variation that increases detection difficulty. Fourth, the occlusion condition varies. Shatian pomelos usually grow in dense clusters, which leads to varying degrees of occlusion and complicates feature extraction from images. To solve these issues in practical applications, a comprehensive multi-strategy framework has been proposed, as shown in Fig. 1. This framework consists of three parts. The first part entails dataset composition tailored to simulate a real-world orchard. Differing from specific scenarios in previous studies, a multi-scenario dataset was constructed through a combination of orchard collection and internet acquisition. Internet acquisition has achieved diversity in captured scenarios and imaging devices. And the collected images specifically address the problem of lighting changes, scale, and occlusion. To further resolve these issues, the second part of the framework involves object image enhancement techniques, which include grayscale transformation, contrast adjustment, and brightness alteration. It also effectively resolves differences in saturation, contrast, and sharpness caused by different imaging devices. The final component of the proposed framework is REAS-Det, a novel detection architecture built upon YOLOv8. To address the inherent limitations of conventional convolutional operators in complex orchard environments, we design a new convolutional module termed Global-Selective Visibility Convolution (GSV-Conv). GSV-Conv addresses the limited visibility of conventional convolution and causal sequence modeling by reconstructing the feature space based on global semantic similarity. Through semantic reordering and prompt-augmented state-space scanning, each spatial location gains access to globally relevant information without relying on bidirectional attention. As a result, convolution is performed over a visibility-aware feature space, achieving non-causal global perception with high computational efficiency. In addition, REAS-Det integrates a dedicated multi-scale feature selection structure to enhance representations of occluded objects, along with a modified Non-Maximum Suppression (NMS) strategy to improve instance separation under dense clustering. Together, these components form a unified and complementary architecture that effectively addresses multi-scale and occlusion challenges in real-world orchard scenarios. This comprehensive solution framework demonstrates significant accuracy improvement in Shatian pomelo detection under real orchard conditions. The specific contributions of this paper are summarized as follows:

(1) To address the challenge of large-scale variation in fruit detection, this study designs a novel convolution module named Global-Selective Visibility Convolution (GSV-Conv). GSV-Conv redefines feature aggregation by selectively constructing a global visibility-aware feature space prior to spatial convolution. This formulation enables scale-adaptive perception across a broad range of object sizes while retaining the computational efficiency and inductive bias inherent to convolutional operations. This design provides a principled and effective solution for detecting multi-scale and densely distributed fruits in real orchard scenarios.

(2) To achieve accurate detection of Shatian pomelos across multiple scales, a Composite Receptive Field Enhancement Module (C3RFEM) is integrated to compensate for the loss of detail during receptive field expansion. To improve the detection accuracy of occlusion, a multi-scale, multi-head feature selection structure is employed. Furthermore, an improved Non-Maximum Suppression algorithm(soft-NMS) is introduced to enhance recognition accuracy.

(3) A multi-strategy fruit detection framework centered on Shatian pomelo is developed, identifying four key variation issues and boosting accuracy through the proposed framework. The proposed method surpasses the advanced models in complexity and accuracy, which is more targeted and effective for Shatian pomelos detection in real orchard environments.

## 2. Materials and methods

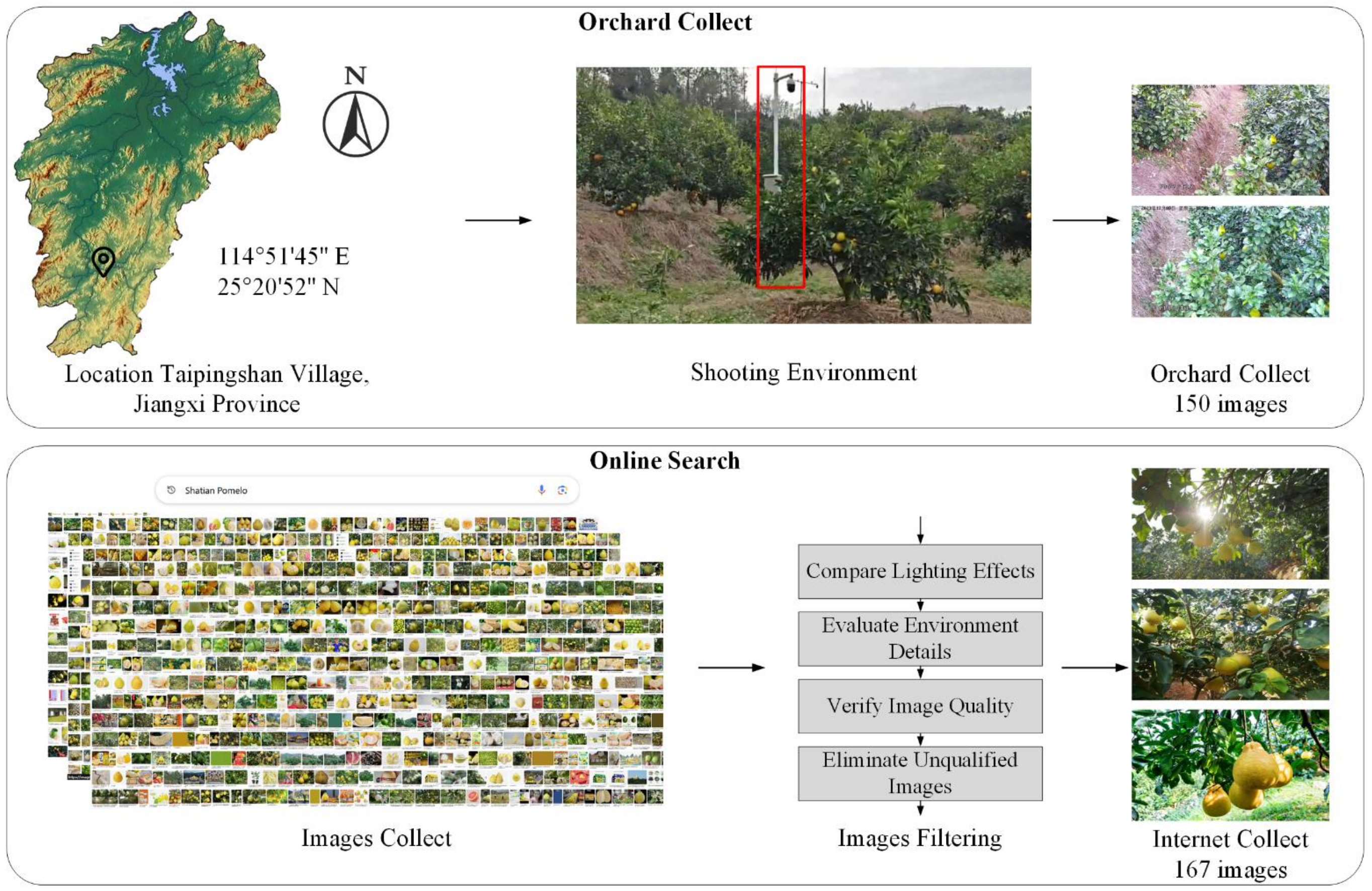

Fig. 2 Image Collection Process. Images were collected from real orchards and online sources.

## 2.1. Image acquisition

As the first part of the framework, mature Shatian pomelos were selected as the research objects. To address four key variation issues, images were collected from real orchards and the internet, as shown in Fig. 2. Preference was given to those with variations in lighting, scale, and occlusion to address the remaining challenges.

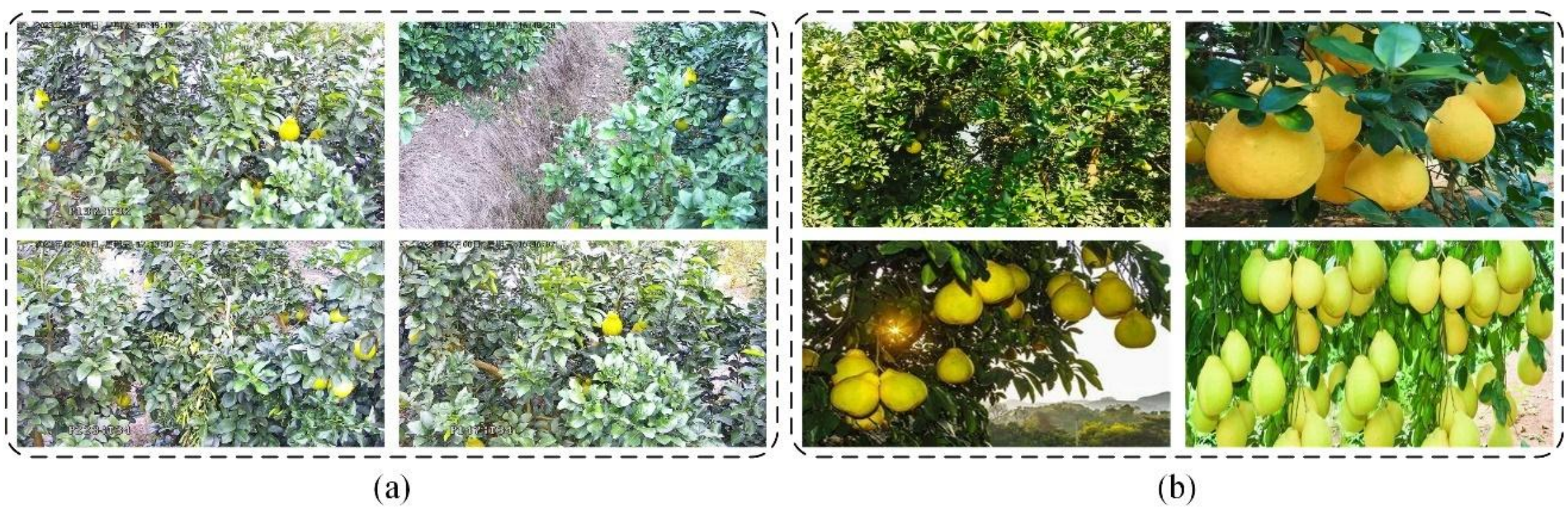

Fig. 3 Example of collected images. (a) Images collected from the orchard. (b) Images collected from the internet.

The first part was taken in December 2023 at an orchard in Taiping Shang Village, Jiading Town, Xinfeng County, China. The imaging device was produced by China's Hikvision Group, model DS-8632N-I8, with a camera resolution of 1920×1080 pixels. After manually removing images with poor pixel quality and blurred images, 150 high-resolution field images were retained for analysis. An example of these images is shown in Fig. 3 (a). The second part was collected from the internet. Under varied lighting conditions, images were selected from different regions with different imaging equipment and planting conditions. The result was that 167 high-quality online images were manually gathered, as shown in Fig. 3 (b). In total, 317 original images selected from the two parts were compiled, which formed the original dataset.

## 2.2. Dataset augmentation and annotation

LabelImg software (https://github.com/HumanSignal/labelImg) was used to annotate Shatian pomelo images. In these images, Shatian pomelos were annotated with rectangles and a single category label "0," and the labels were saved in the YOLO "xml" format. Following the annotation, the limited scale of the original dataset necessitated enhanced image diversity to improve model detection performance for Shatian pomelos under different environmental conditions. To mitigate insufficient training data impacts on model effectiveness and generalization, data augmentation techniques were implemented. These included random flipping, grayscale transformation, noise addition, contrast adjustment, and brightness modification, collectively creating various transformed images to enrich the training sample set. Moreover, 190 original images from the dataset were randomly selected for a 7-fold single expansion, resulting in 1330 training images. Examples of the enhancement effects are shown in Fig. 4. The remaining 127 original images were used as the validation sets for the model.

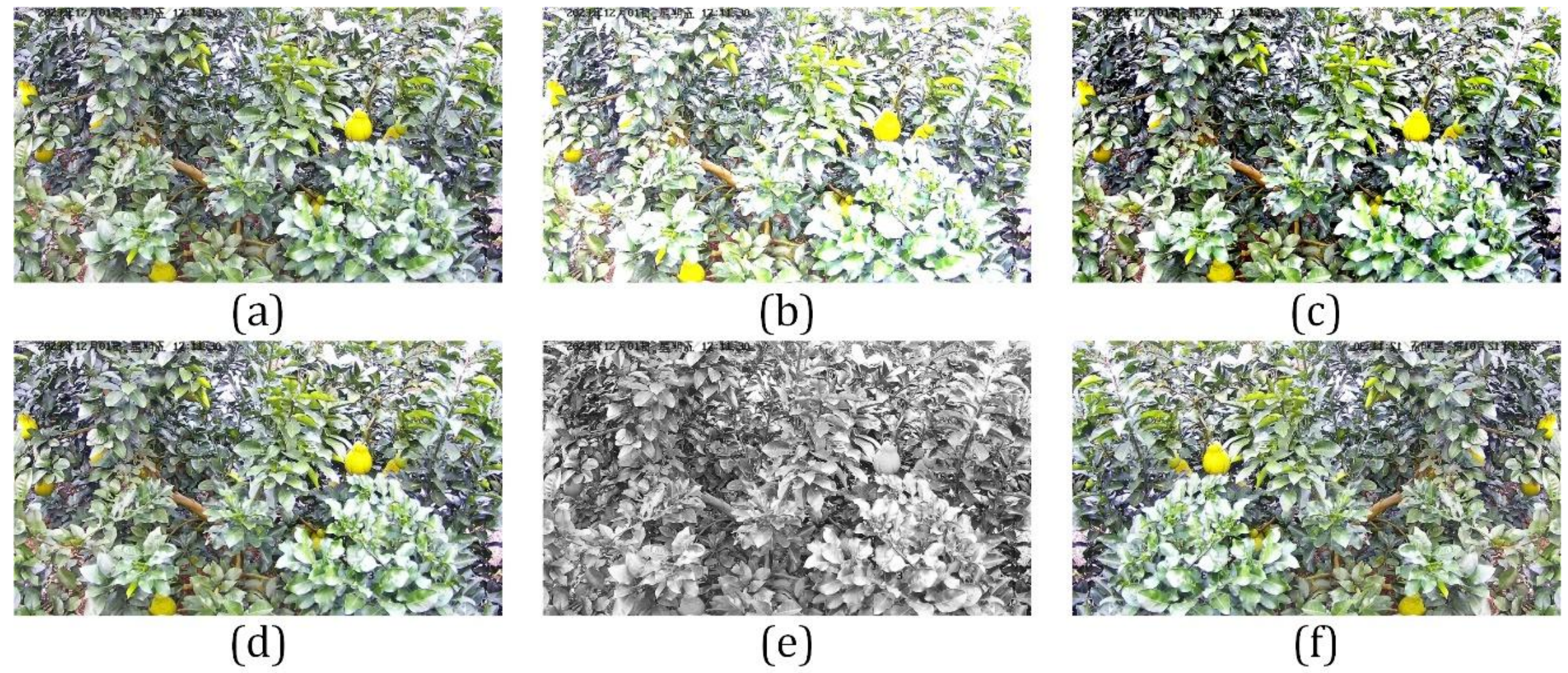

Fig. 4 Examples of Enhancement Effects. (a) Original image, (b) Increased brightness, (c) Increased contrast, (d) Added noise, (e) Grayscale transformation, (f) Horizontal flip

## 2.3. Global-Selective Visibility Convolution (GSV-Conv)

Even when equipped with multi-scale designs or attention mechanisms, conventional convolution fundamentally performs spatial aggregation within a local and predefined receptive field. Moreover, when models involve sequential modeling or linear state-space formulations, additional causal constraints are imposed, which render many potentially useful long-range features invisible during representation learning.

To address these limitations, we propose Global-Selective Visibility Convolution (GSV-Conv). Traditional attention reweights visible features, whereas our method redefines which features are visible before any spatial aggregation takes place. Rather than introducing another attention mechanism, our approach focuses on resolving the problem of insufficient feature visibility at its source.

Specifically, we employ a non-causal global modeling mechanism that endows features with global semantic visibility prior to aggregation, enabling spatially distant yet semantically related regions to be perceived simultaneously. On this basis, adaptive selection of spatial receptive ranges is further introduced, allowing the model to dynamically determine the appropriate aggregation scale according to local structural complexity and contextual requirements. Consequently, convolution no longer operates passively within a constrained field of view, but instead performs selective aggregation over a semantically reconstructed visible space.

In terms of effectiveness, this design achieves a favorable balance between efficiency and representational capacity. On the one hand, it preserves the efficiency and stability of convolution for local modeling, avoiding the high computational cost associated with global attention mechanisms. On the other hand, it substantially enhances the model's ability to capture long-range dependencies, small objects, and complex backgrounds, with particularly pronounced advantages in scenarios where semantically related regions are spatially dispersed. Accordingly, we term this approach *visibility-driven convolution*, which goes beyond naive receptive-field enlargement by first enhancing local receptive-field representations and subsequently expanding the domain of visible features for fine-grained spatial selection within a reconstructed visibility space, as illustrated in Fig. 5. The entire convolution process can be decomposed into three sequential stages:

Fig.5 Architecture of the proposed GSV-Conv module.

### 2.3.1. Global-Selective Visibility Construction

First, given an input feature map $\Phi \in \mathbb{R}^{C\times H\times W}$, we first represent it as a sequence

$$\mathbf{S} = \mathrm{Flatten}(\Phi) \in \mathbb{R}^{N\times C}, \quad N = H\times W \tag{1}$$

where each element $\mathbf{s}_i$ corresponds to a spatial location in the original feature map. Rather than modeling this sequence according to the raster-scan order, we introduce a semantic routing mechanism to determine feature visibility at a global level.

Specifically, a lightweight routing function $f_r(\cdot)$ predicts a categorical distribution over $K$ latent semantic tokens for each sequence element,

$$\mathbf{p}_i = f_r(\mathbf{s}_i), \quad \mathbf{p}_i \in \mathbb{R}^{K} \tag{2}$$

A discrete routing assignment is obtained via Gumbel-Softmax sampling,

$$z_i = onehot\left(\arg\max_{k\in\{1,\dots,K\}} \frac{\exp((\log p_{i,k}+g_{i,k})/\tau)}{\sum_{j=1}^{K}\exp((\log p_{i,j}+g_{i,j})/\tau)}\right), \quad g_{i,k} = -\log(-\log u_{i,k}), u_{i,k} \sim Uniform(0,1) \tag{3}$$

which assigns each feature to a semantic group without performing aggregation. These assignments define semantic visibility, indicating which features should be considered semantically related regardless of spatial distance. Based on the semantic routing assignments $\mathbf{Z}$, global-selective visibility construction jointly performs semantic reordering and global context encoding. Specifically, the reordered sequence and the corresponding token-conditioned prompts are obtained as

$$(\mathbf{S}', \mathbf{P}) = \left(\pi(\mathbf{S}, \mathbf{Z}), \mathbf{ZE}\right) \tag{4}$$

where $\pi(\cdot)$denotes a permutation operator that reorganizes the sequence according to semantic assignments, and $\mathbf{E} \in \mathbb{R}^{K\times d}$is a learnable semantic token embedding matrix. The permutation operation explicitly breaks the original spatial ordering and groups spatially distant but semantically related features into adjacent positions, thereby reconstructing semantic visibility at the sequence level. Meanwhile, the generated prompts $\mathbf{P}$ encode global semantic context associated with each token and are used to modulate the subsequent sequence modeling process, enabling global semantic information to influence feature evolution during scanning.

### 2.3.2. Adaptive Receptive-Field Aggregation

Given an input feature Φ, receptive-field spatial aggregation is formulated as

$$\mathcal{F}(\Phi) = A_{rf} \times \max\left(0, \mathcal{N}(W_k * \Phi + b_k)\right) \tag{5}$$

$$A_{rf} = \frac{e^{W_1 \cdot \phi(\Phi) + b_1}}{\sum_{j=1}^{C} e^{(W_1 \cdot \phi(\Phi) + b_1)_j}} \tag{6}$$

where $A_{rf}$ is the attention map and $F_{rf} = \max\left(0, \mathcal{N}(W_k * \Phi + b_k)\right)$ is the transformed receptive fields spatial features. Here, $W_1$ and $W_k$ represent learnable parameters corresponding to the $1 \times 1$ and $k \times k$ convolutions, $\mathcal{N}(\cdot)$ represents normalization, $\Phi$ represents the input feature map, and $\mathcal{F}(\Phi)$ is obtained by multiplying the attention map $A_{rf}$ with the transformed receptive fields spatial features $F_{rf}$.

### 2.3.3. Non-Causal Semantic Sequence Refinement

The visibility-aware sequence $\mathbf{S}'$, together with the token-conditioned prompts $\mathbf{p}_i^{(g)}$, is processed by a selective scan state-space model. At each sequence position $t$, the hidden state is updated as

$$\mathbf{h}_t = \mathbf{A}_t \mathbf{h}_{t-1} + \mathbf{B}_t \mathbf{s}_t' \tag{7}$$

and the output is computed via

$$\mathbf{y}_t = (\mathbf{C}_t + \mathbf{p}_t^{(g)}) \mathbf{h}_t + \mathbf{D}\mathbf{s}_t' \tag{8}$$

where $\mathbf{A}_t, \mathbf{B}_t, \mathbf{C}_t$ are input-dependent parameters. By injecting global semantic prompts into the output projection, non-causal global semantic perception is achieved without modifying the causal state transition.

After inverse reordering and reshaping back to the spatial domain, the sequence modeling output is integrated with the locally aggregated feature through residual refinement,

$$\mathbf{Y}_{out} = \mathbf{Y}_{SSM} + \mathbf{X}_{RF} \tag{9}$$

where $\mathbf{X}_{\mathrm{RF}}$ denotes the receptive-field aggregated feature map. This design allows global semantic information to refine locally consistent representations while preserving spatial coherence.

GSV-Conv differs fundamentally from attention-based convolutional variants. Rather than reweighting features that are already visible, GSV-Conv redefines which features are visible prior to global modeling. By jointly performing global-selective visibility construction and receptive-field spatial aggregation, GSV-Conv enables efficient non-causal global semantic interaction while retaining the robustness and efficiency of convolution.

## 2.4. Improvement of the YOLOv8 model

In real orchards, high planting density creates significant detection challenges. Single images frequently contain fruits at heterogeneous distances (far-to-near), introducing substantial scale variation that degrades detection performance. Additionally, the natural growth pattern of causes mutual occlusion, severely impeding visual feature extraction. To accurately identify Shatian pomelos in real orchards, this paper proposes the REAS-Det network, with its architecture illustrated in Fig. 6. In the backbone network, the GSV-Conv module incorporating global-selective visibility modeling is first introduced. This convolution operation selectively expands the effective visible feature space and enhances feature extraction capability. Subsequently, the C3RFEM module, designed based on the RFE receptive fields enhancement module, replaces the traditional C2f operation. It retains convolution operations during upsampling to better capture image details and merge receptive fields. The architecture further integrates a MultiSEAM feature selection network at the head layer to mitigate occlusion effects, while substituting the original NMS with the soft-NMS for enhanced localization precision. These enhancements in the REAS-Det model effectively improve detection accuracy and robustness, enabling reliable performance in complex orchard environments characterized by dense foliage and variable lighting.

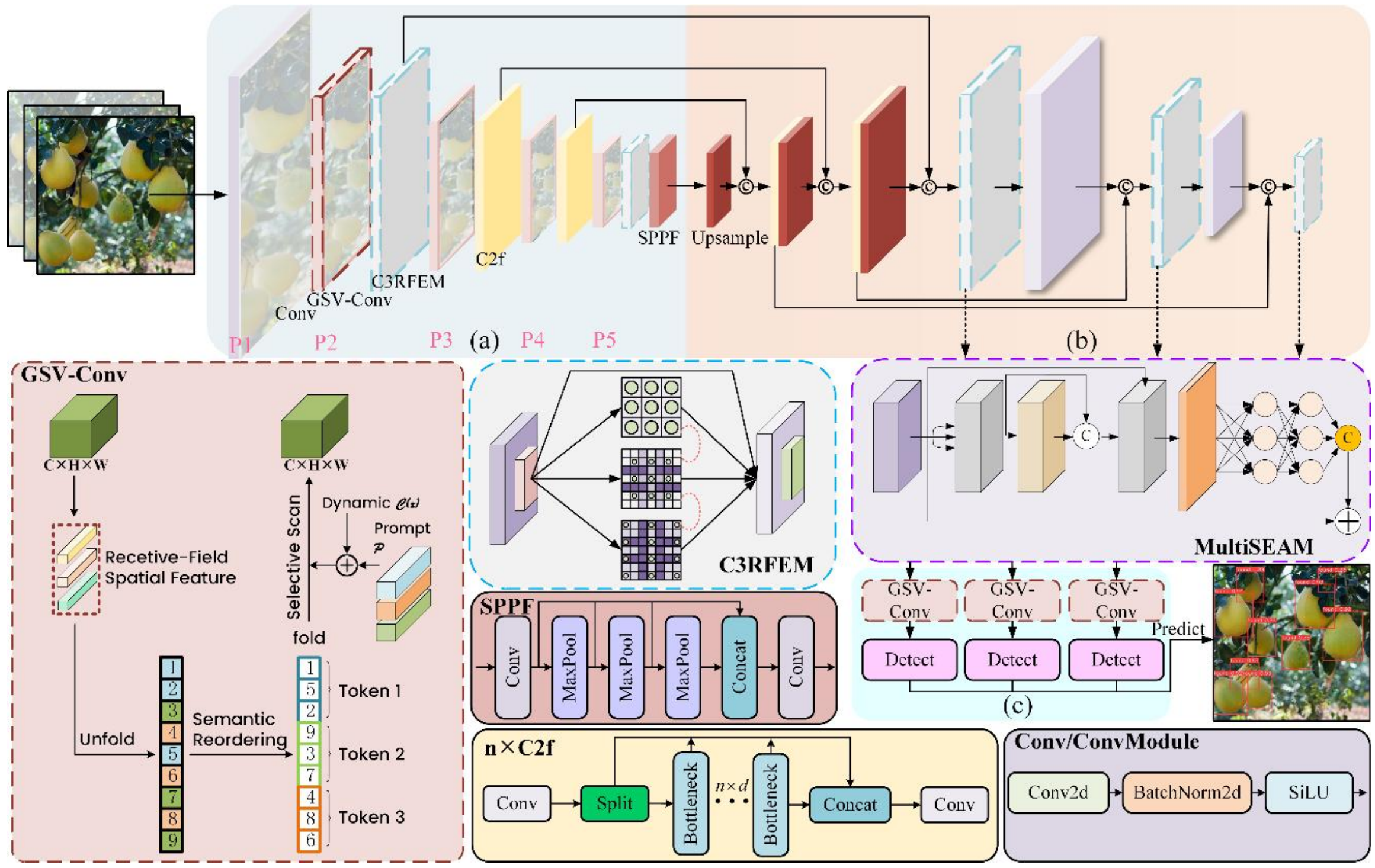

Fig. 6 The REAS-Det network architecture. It consists of: (a) Backbone: The original Conv layers have been replaced by GSV-Conv layers that enlarge the receptive fields. The C2f modules in layers P3 and P5 have been substituted by Receptive fields Enhancement Modules (C3RFEM). Spatial Pyramid Pooling (SPPF) segregates the most critical contextual features and enlarges the receptive fields. (b) Neck: Composed of multiple convolutional and pooling layers. (c) Head: The feature vectors processed by the MultiSEAM module undergo GSV-Conv operations to produce the final detection outputs.

### 2.4.1. Receptive Fields Expansion (RFE) Module

Following the identification of receptive fields' key features in the RFAConv module, a Scale-Aware Receptive Field Expansion (RFE) (Yu, et al., 2024) module is introduced. This module leverages dilated convolutions to expand the receptive field, enhancing the feature maps' ability to capture contextual information. Its four branches with varying dilation rates extract multi-scale features and recognize long-range dependencies. These branches share weights and have different receptive fields, which is useful to reduce model parameters and mitigate the risk of overfitting, effectively utilizing each sample. As shown in Fig. 7, this module consists of two key components: the multi-dilation convolution branches and the aggregation weighting layer. The former uses a fixed 3x3 convolution kernel with dilation rates of 1, 2, and 3; the latter integrates information from various branches and applies weighting to balance the representation of features from each branch. Additionally, to prevent potential gradient issues during training, a residual connection was added in this module. To better assimilate features, a traditional C3 module, which retains convolution operations during the upsampling process, is integrated instead of the C2f module. The C3RFEM was constructed by combining this RFE module with the C3 module, while replacing select C2f modules in YOLOv8, which enhanced the detection and recognition accuracy of Shatian pomelos across different sizes.

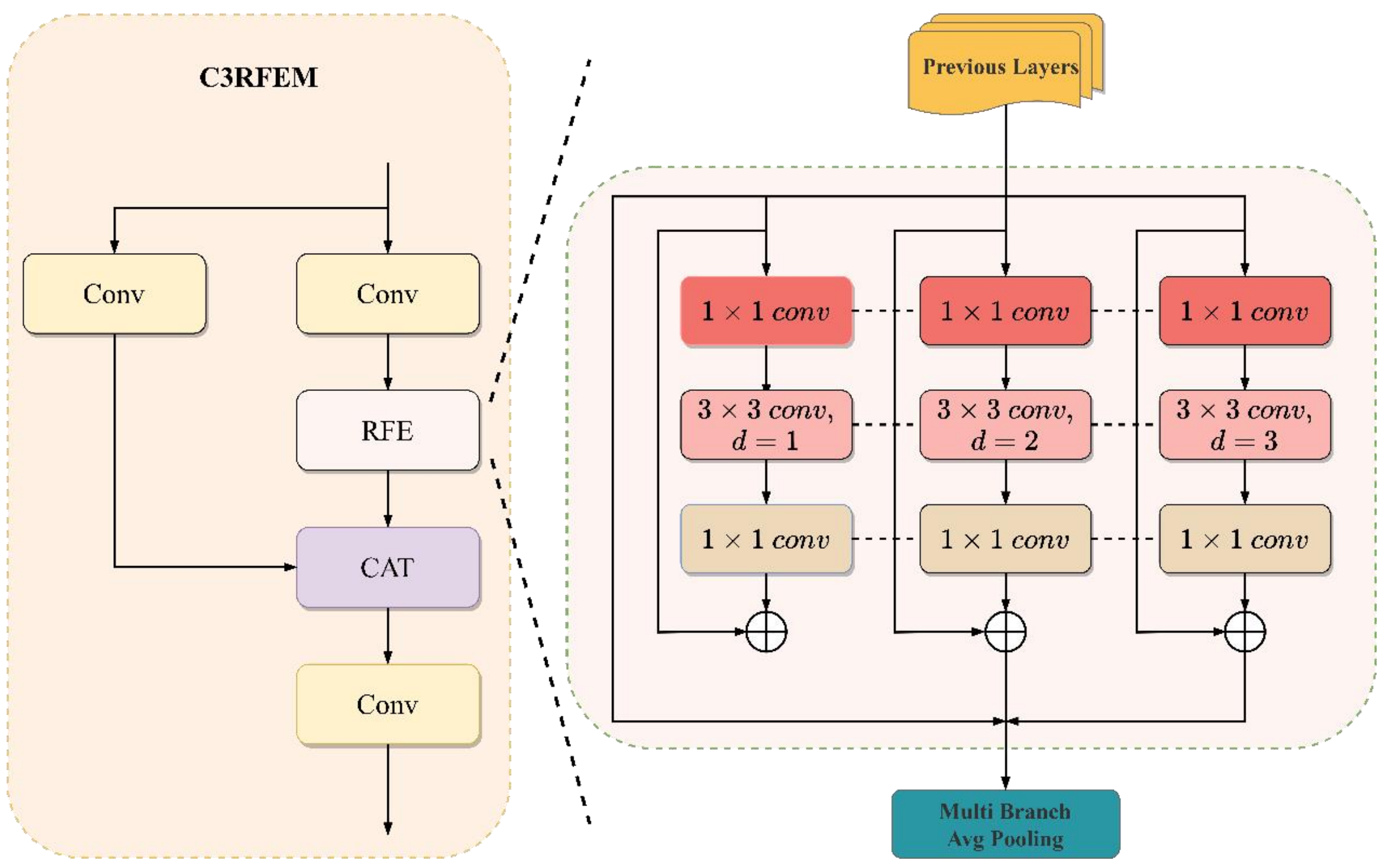

Fig. 7 The detailed structure of C3RFEM. It integrates a Scale-Aware Receptive Fields Expansion (RFE) with multi-branch dilated convolutions and a C3 module to enhance multi-scale feature representation and improve pomelo detection accuracy.

### 2.4.2. MultiSEAM feature selection structure

To address the issue of occlusion recognition in Shatian pomelos more effectively, a new multi-scale, multi-head feature selection structure at the top layers of this model, named MultiSEAM (Yu, et al., 2024), was introduced. MultiSEAM is designed as a lightweight, occlusion-aware channel–attention block that can be seamlessly plugged into any convolutional backbone. Given an input feature map $\Phi \in ð^{C \times H \times W}$, the module operates in three sequential stages(as shown in Fig. 8). First, a depthwise separable convolution is applied to retain spatial locality while reducing parameter overhead. A point-wise $1 \times 1$ convolution then mixes information across channels, and a residual shortcut preserves the original representation, yielding

$$F_{c,h,w}^{dw} = \sum_{i=1}^{k}\sum_{j=1}^{k} W_{c,i,j}^{dw} \Phi_{c,h+i-p-1,w+j-p-1}, \quad 1 \le h \le H, 1 \le w \le W \tag{10}$$

$$F_{1,c,h,w} = \sum_{c=1}^{C} W_{\hat{c},c}^{pw} \, F_{c,h,w}^{dw} + \Phi_{c,h,w}, \quad \hat{c} = 1, \ldots, C \tag{11}$$

where $c \in \{1, \ldots, C\}$ and $p = (k-1)/2$. Second, channel-wise importance scores are inferred from $F_{1,c,h,w}$. Global average pooling aggregates spatial information; the resulting vector passes through a two-layer fully connected bottleneck with ReLU activation, after which a sigmoid squashes the values to [0, 1]. To enlarge the dynamic range of the weights, an exponential mapping shifts the interval to [1, e]:

$$s = \sigma\Big(\sum_{d=1}^{C/r} W_{c,d}^{(2)} \Big[\sum_{c'=1}^{C} W_{d,c'}^{(1)} \frac{1}{HW} \sum_{h',w'} F_{1,c',h',w'}\Big]_{+}\Big) \tag{12}$$

$$A_c = \exp(s), \quad A_c \in \eth^{C\times1\times1} \tag{13}$$

where $[x]_+ = max\{0, x\}$ is the rectified linear (ReLU) operator expressed as a pure max. Finally, the learned attention vector $A_c$ is broadcast along spatial dimensions and applied to the original feature map through channel-wise multiplication, producing the refined output

$$\mathrm{F}_{c,h,w} = A_c \odot \Phi_{c,h,w} \quad (c = 1, \ldots, C; h = 1, \ldots, H; w = 1, \ldots, W) \tag{14}$$

where $\odot$ denotes element-wise multiplication with automatic broadcasting.

By coupling depthwise separable convolutions with exponentially scaled channel attention, MultiSEAM effectively enhances this model's ability to recognize Shatian pomelos against complex backgrounds while improving its adaptability and accuracy in occluded Shatian pomelos' recognition.

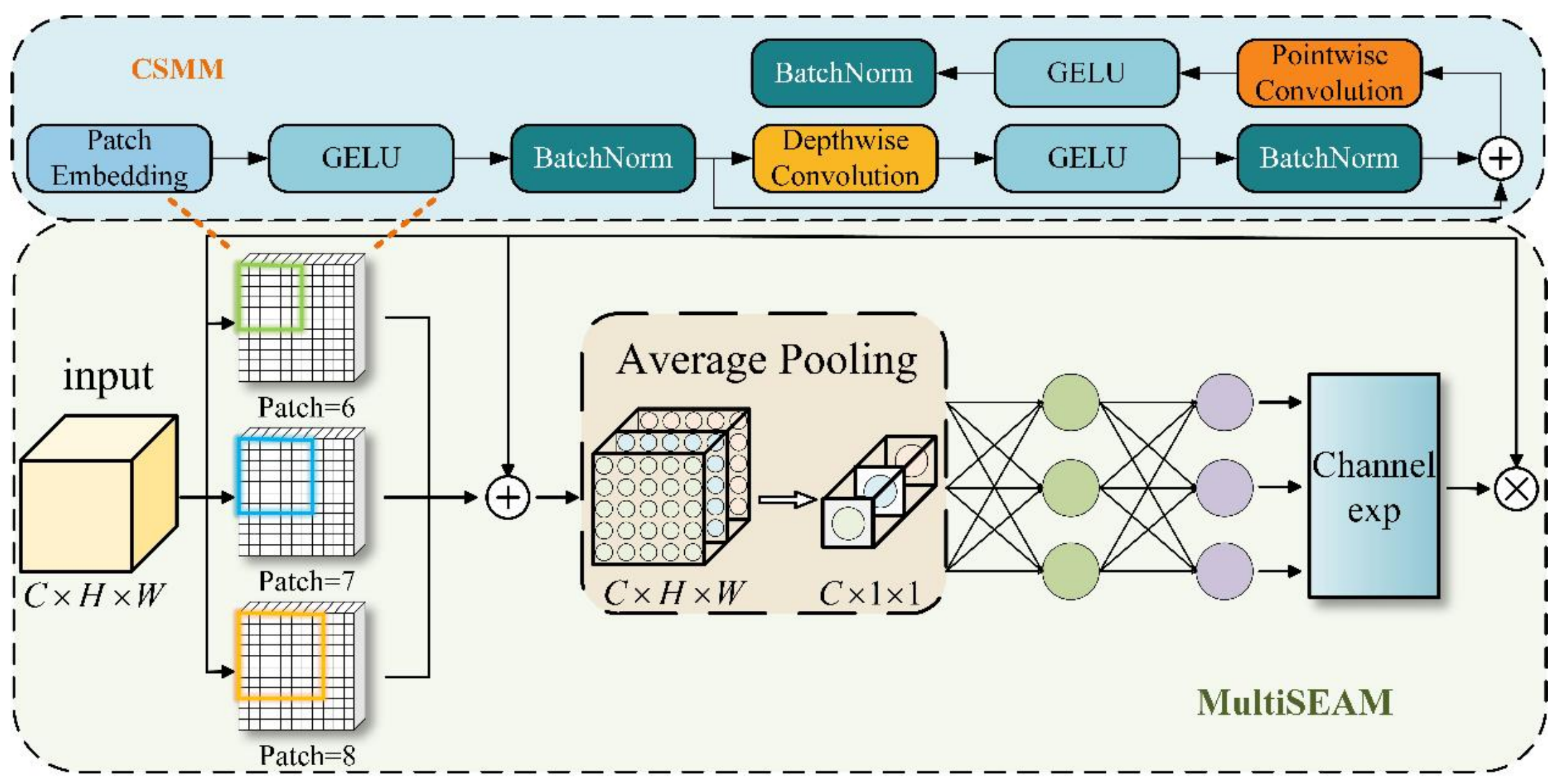

Fig. 8 Architecture of the MultiSEAM Module. The depicted structure outlines the architecture of the MultiSEAM module, with the upper section representing the Channel and Spatial Mixing Module (CSMM). CSMM processes multi-scale features through patches and employs depthwise separable convolutions to capture spatial and channel correlations.

### 2.4.3. soft-NMS

In YOLOv8, the non-maximum suppression (NMS) algorithm is used to remove duplicate bounding boxes. However, this method can mistakenly suppress detection boxes that overlap but represent different objects, particularly in scenes with dense objects. This problem becomes pronounced when detecting Shatian pomelos, as most cover small pixel areas in images, especially when branches are dense and fruit quantity is high, resulting in significant fruit overlap. To avoid missing or double-counting, this study employs the soft-NMS (Bodla, Singh, Chellappa, & Davis, 2017) to enhance model performance. The core idea is that when the overlap ratio of two anchor boxes exceeds $\eta_0$ , their scores are not set to zero directly. Instead, a Gaussian function is used to decay the score of the anchor box, as shown in Equation 4.

$$\mathrm{s}_i = s_i e^{-\frac{\eta(b^*, b_i)^2}{\sigma}} \tag{15}$$

where $s_i$ represents the score of the *i*-th detected object, $\eta(b^*, b_i)$ denotes the overlap ratio between two anchor boxes, and $\sigma$ is an adjustable parameter that controls the speed and range of the score decay. As observed from this equation, the higher the overlap ratio, the greater the score reduction. And when the degree of overlap between two boxes increases, the corresponding box's score decreases more rapidly, thereby significantly improving the accuracy of detection boxes. The complete soft-NMS algorithm is detailed in Alg. 1.

**Alg. 1 Pseudocode of the soft-NMS algorithm**

**Input:** initial detection box set $\mathcal{B}$, detection scores set $\mathcal{S}$, NMS threshold $\eta_0$

**Begin**

- $\mathcal{B} \leftarrow \{b_i\}, \mathcal{S} \leftarrow \{s_i\}, \eta(b_u, b_v) \leftarrow \eta_0$
- $\mathcal{B}' = \mathcal{S}' = \{\}$ // $\mathcal{B}', \mathcal{S}'$ are detection box set and detection scores set
- **While** $\mathcal{B} \neq \emptyset$ do
  - $s^* \leftarrow argmax\ \{s_i\}$
  - $b^* \leftarrow s^*$
  - $\mathcal{B}' \leftarrow \mathcal{B}' \cup b^*$
  - $\mathcal{S}' \leftarrow \mathcal{S}' \cup s^*$
  - $\mathcal{B} \leftarrow \mathcal{B} - b^*$
  - $\mathcal{S} \leftarrow \mathcal{B}$
  - for $b_i \in \mathcal{B}$
    - $$s_i \leftarrow \begin{cases} s_i & \eta(b^*,\ b_i) \leq \eta_0 \\ s_i(1 - \eta(b^*,\ b_i)) & \eta(b^*,\ b_i) \geq \eta_0 \end{cases}$$
  - end
- **end**
- return $\mathcal{B}', \mathcal{S}'$

**end**

## 2.5. Model training

In this study, to maintain experimental rigor, all models were trained under identical conditions. For ablation studies, all hyperparameters were kept as default, while model comparison experiments used each model's default hyperparameter configuration. Additionally, no hyperparameter optimization was performed, nor were hyperparameters adjusted based on validation set results. Given these conditions, the validation set can be independently used to directly evaluate model performance, allowing more images can be included and providing a broader range of data samples.

The experimental setup for this study is based on Ubuntu 22.04 OS, equipped with 120 GB RAM and a single NVIDIA RTX-4090 GPU. The deep learning framework deployed is PyTorch 2.2.1, utilizing CUDA 12.1 for computational acceleration. In this study, the default AdamW optimizer was employed to fine-tune model parameters, with an initial learning rate of 0.01. The training process included a maximum of 300 epochs. Due to the early stopping mechanism in the YOLOv8 model, training was halted early if the model's validation performance ceased to improve, thereby preventing overfitting. As a result, some experiments did not reach the full 300 epochs. The input image size was set to 640 × 640, with a batch size of 8 images per training iteration, and the number of workers was set to 8. The source code for the REAS-Det will be available at https://github.com/Genk641/REAS-Det.

## 2.6. Model evaluation metrics

Object detection models need to be evaluated based on several metrics. To assess the performance of REAS-Det, four evaluation metrics were used: precision, recall, mAP@.50 (mean Average Precision), and mAP@.50:.95. The formulas for calculating these performance metrics are shown below:

$$\mathrm{P} = \frac{TP}{TP + FP} \tag{16}$$

$$R = \frac{TP}{TP + FN} \tag{17}$$

$$AP = \int_0^1 P(R)dR \tag{18}$$

$$mAP = \frac{AP_1 + AP_2 + \cdots + AP_n}{n} \tag{19}$$

$$mAP@.50:.95=\frac{1}{10}\sum_{r=10}^{19}mAP@\frac{r}{20} \quad (20)$$

where TP represents the number of true positives, which are samples correctly classified as positive; FN stands for the number of false negatives, referring to positive samples incorrectly classified as negative; FP denotes the number of false positives, which are negative samples mistakenly classified as positive; TN indicates the number of true negatives, meaning samples that were correctly classified as negative. The variable n represents the number of classes. This demonstrates that precision serves as a valid metric for evaluating this model's accuracy in predicting the target. High precision indicates fewer false positives in the model's outputs. When precision is high, detections classified as target objects demonstrate high reliability, whereas recall quantifies the model's ability to capture true positive cases, where a high recall rate indicates robust detection of positive cases and minimized false negatives. The metric mAP@.50 calculates the mean Average Precision when the Intersection over Union (IoU) threshold is set to 0.50. IoU represents the overlap ratio between the predicted under a more lenient IoU threshold. Conversely, mAP@.50:.95 is the average mAP calculated over a range of IoU thresholds from 0.50 to 0.95. mAP@.50:.95evaluates this model's stability and accuracy when higher localization precision is required. In summary, these four evaluation metrics provide a comprehensive assessment of the REAS-Det model's performance in detecting Shatian pomelos under complex orchard conditions.

## 3. Results and discussion

### 3.1 Comparison with Different Convolution Modules

To verify the effectiveness and generalization capability of the proposed convolution module in object detection tasks, we conduct comprehensive comparative experiments within a unified YOLOv8 framework. To ensure fairness and reproducibility, the overall network architecture, training strategy, hyperparameter settings, and data split are strictly kept identical across all experiments, and only the basic convolutional units in YOLOv8 are replaced. Specifically, depthwise separable convolution (DWConv), DCNv4, and several other representative advanced convolution variants are adopted for comparison. The quantitative results are summarized in Table 1.

**Table 1. Experimental Results of Comparison with Different Convolution Modules**

| convolution module | Proposed year | Parameters(M) | GFLOPs(↓) | mAP@.50 | mAP@.50:.95 |
| --- | --- | --- | --- | --- | --- |

| SPDConv (Sunkara & Luo, 2022) | 2022 | 3.0 | 8.9 | 78.5 | 45.9 |
|---|---|---|---|---|---|
| ODConv (C. Li, Zhou, & Yao, 2022) | 2022 | 3.0 | 8.0 | 79.0 | 45.9 |
| DySnakeConv (Y. Qi, He, Qi, Zhang, & Yang, 2023) | 2023 | 3.1 | 33.9 | 79.5 | 46.1 |
| SCConv (J. Li, Wen, & He, 2023) | 2023 | 3.0 | 8.6 | 79.8 | 46.3 |
| ARConv (X. Wang, Zheng, Shao, Duan, & Deng, 2025) | 2025 | 3.2 | **1.5** | 27.5 | 15.7 |
| FCB (H. Sun, et al., 2025) | 2025 | 4.4 | 10.8 | 78.5 | 46.3 |
| FDConv (L. Chen, Gu, Li, Yan, & Fu, 2025) | 2025 | 3.0 | 7.9 | 78.9 | 46.0 |
| GSV-Conv (ours) | - | **3.0** | 8.6 | **80.9** | **46.5** |

Experimental results demonstrate that, under the same computational budget, incorporating the proposed convolution module consistently yields notable improvements in detection accuracy. Moreover, compared with other convolutional alternatives, the proposed method achieves a more favorable accuracy–efficiency trade-off while maintaining controllable increases in parameter count and computational complexity, thereby substantiating the effectiveness and novelty of GSC-Conv.

### 3.1. Ablation experiments

Ablation studies were systematically performed to validate REAS-Det's efficacy in Shatian pomelo detection. Based on the YOLOv8 model, each proposed enhancement, individually and collectively, was incrementally incorporated and evaluated in the REAS-Det model. The hardware environment and hyperparameter settings for training all models were kept consistent throughout the experiments. Fig. 9 presents the ablation experiment results for mAP@.50 and mAP@.50:.95, showing the performance of different improvement methods in terms of these metrics. Due to the early stopping mechanism in YOLOv8, training stopped if the validation set's evaluation accuracy and loss values did not show significant improvement, which effectively prevented overfitting before reaching the preset epoch limit. Therefore, the training epochs for different methods may vary. It can be observed that the mAP@.50 and mAP@.50:.95 of the REAS-Det model are higher than those of the original YOLOv8n and other individual improvement methods.

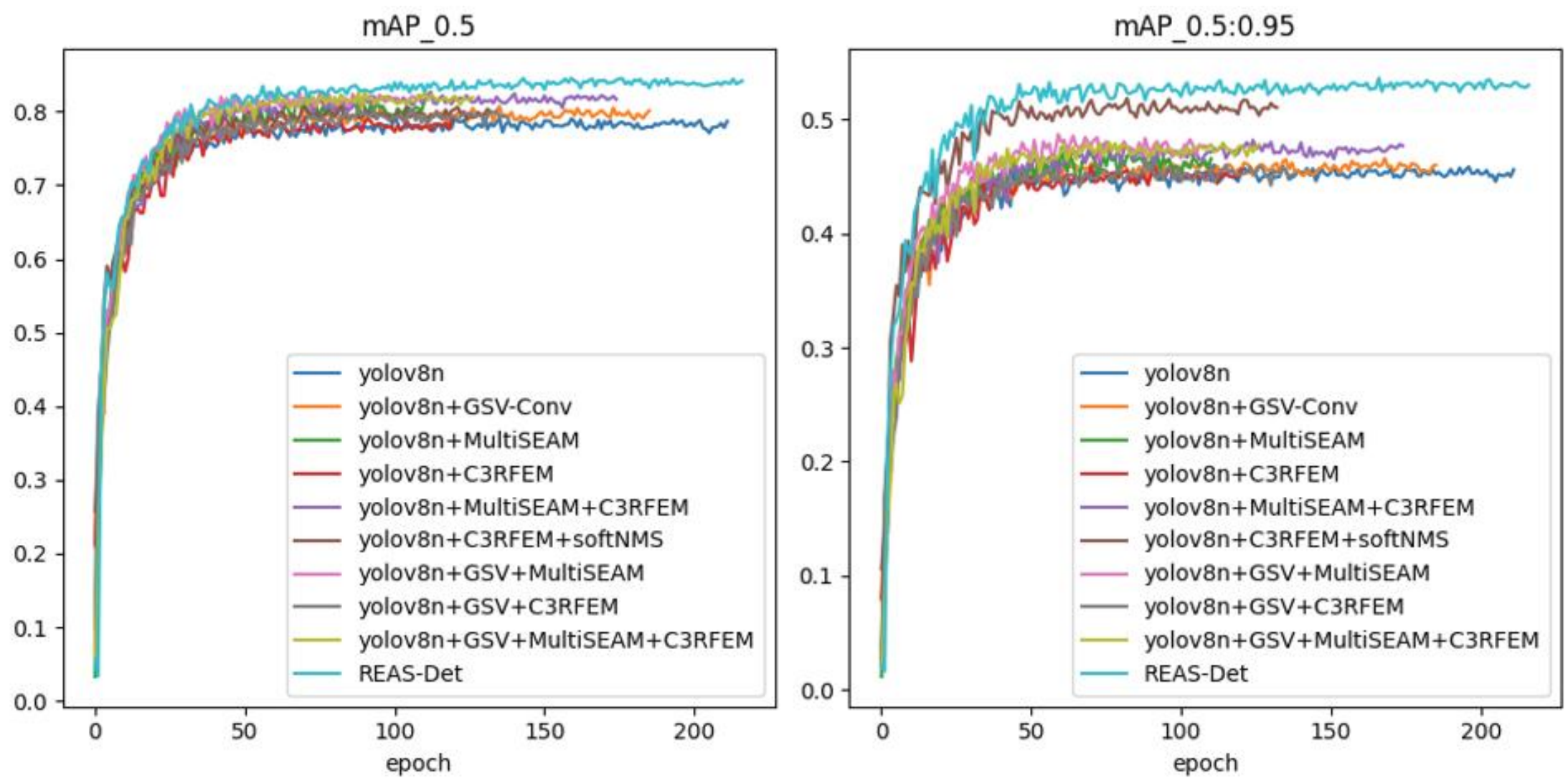

Fig. 9 Ablation results on mAP@.50 and mAP@.50:.95. The proposed REAS-Det model outperforms all baseline variants, demonstrating superior convergence and accuracy in Shatian pomelo detection.

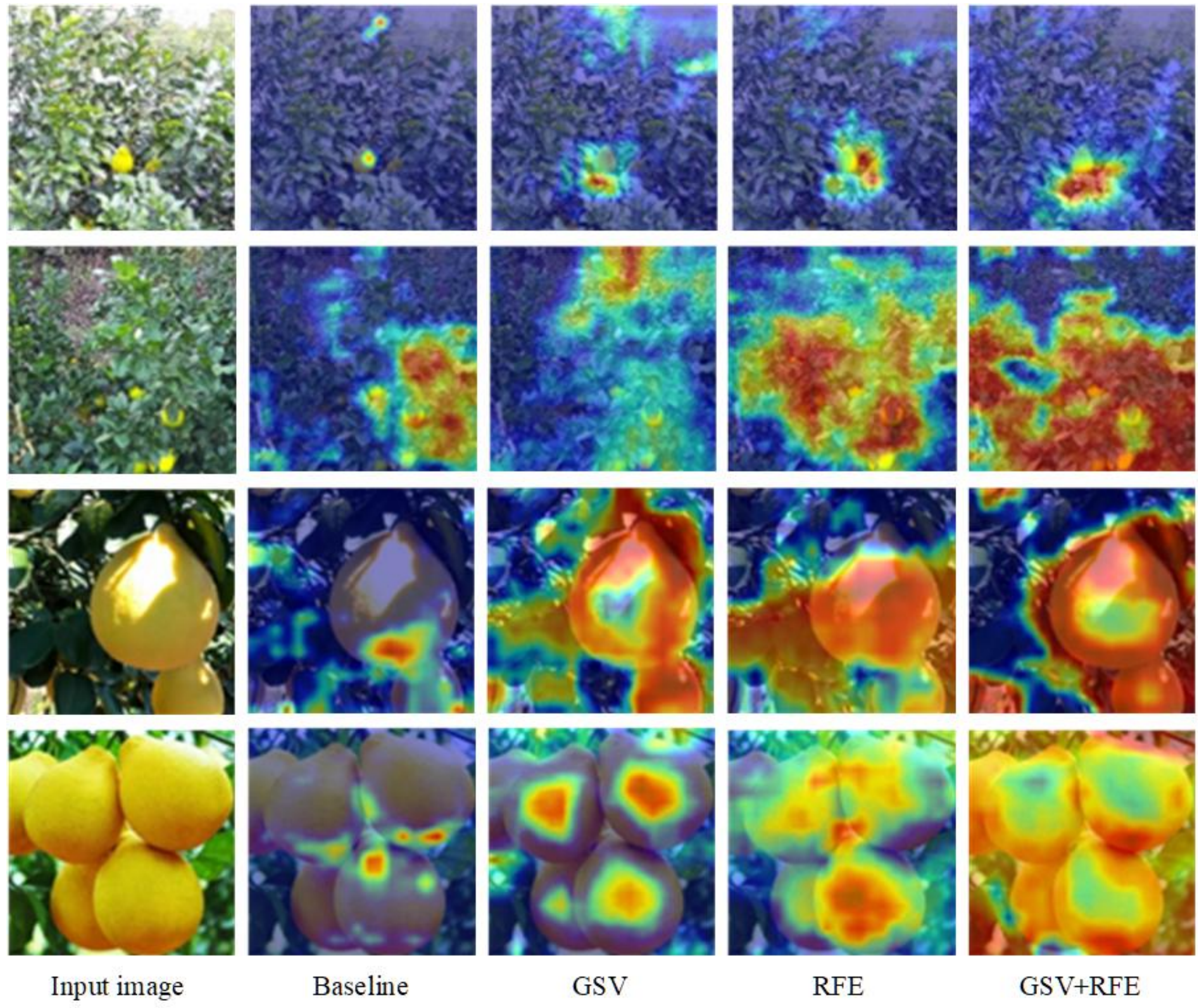

Fig. 10 Visualization of feature maps with GSV and RFE modules. The combined module highlights pomelo regions more clearly,

indicating improved feature localization and representation.

To evaluate the effectiveness of the GSV and RFE modules in enhancing receptive fields, feature maps generated by the proposed methods were analyzed. As illustrated in Fig. 10, the incorporation of the GSV and RFE modules enables the model to focus more on object-relevant regions, thereby capturing richer and more discriminative feature details.

Table 2 compares the ablation experiment results of the improved YOLOv8n model and the REAS-Det model. Due to the large number of added modules, and some requiring combinations to achieve optimal performance, only the effective combinations are listed. The results show that while the model incorporating the C3RFEM and soft-NMS modules has a high precision rate, its recall is relatively low, leading to many missed detections of Shatian pomelos, which would not meet the requirements of practical applications. In contrast, the REAS-Det model achieves mAP@.50 and mAP@.50:.95 of 84.3% and 53.6%, respectively, representing improvements of 6% and 7.6% over the baseline YOLOv8n model. It also maintains the highest precision and a high recall rate. This ablation experiment demonstrates the effectiveness of improved modules for the Shatian pomelo detection task.

**Table 2. The result of the ablation experiments**

| Methods | GSV-Conv | Multi SEAM | C3RFEM | soft-NMS | P(%) | R(%) | mAP @.50 | mAP @.50:.95 |
|---|---|---|---|---|---|---|---|---|
| YOLOv8n | | | | | 84.5 | 70.2 | 78.3 | 46.0 |
| YOLOv8n+GSC-Conv | √ | | | | 85.7 | 73.3 | 80.9 | 46.5 |
| YOLOv8n+MultiSEAM | | √ | | | 84.8 | 72.3 | 80.4 | 47.6 |
| YOLOv8n+ C3RFEM | | | √ | | 84.9 | 71.5 | 79.9 | 51.6 |
| YOLOv8n+ MultiSEAM+C3RFEM | | √ | √ | | 85.4 | 75.6 | 82.2 | 48.2 |
| YOLOv8n+C3RFEM+soft-NMS | | | √ | √ | **87.2** | 70.5 | 81.0 | 51.9 |
| YOLOv8n+GSV--Conv +MultiSEAM | √ | √ | | | 85.2 | 74.9 | 82.6 | 48.6 |
| YOLOv8n+GSV--Conv +C3RFEM | √ | | √ | | 85.0 | 74.7 | 79.8 | 46.3 |
| YOLOv8n+GSV-Conv +MultiSEAM+C3RFEM | √ | √ | √ | | 82.9 | 76.0 | 81.9 | 48.1 |
| REAS-Det | √ | √ | √ | √ | 86.5 | **77.2** | **84.3** | **53.6** |

## 3.2. Comparative Experiments

To further validate the performance advantages of the proposed model, REAS-Det was evaluated with mainstream object detection networks, including Faster R-CNN (Ren, He, Girshick, & Sun, 2015), RTMDet (C. Lyu, et al., 2022), Cascade R-CNN (Cai & Vasconcelos, 2019), DDQ-4scale (S. Zhang, et al., 2023), YOLOv5 (Jocher, et al., 2022), and YOLOv9 (C.-Y. Wang, Yeh, & Liao, 2024). The results are shown in Table 3. It can be observed that this model outperforms the others in terms of accuracy. Specifically, REAS-Det achieved a 30.6% higher mAP@.50 and a 33.4% higher mAP@.50:.95 compared to Faster R-CNN. Relative to Rtmdet, it showed an improvement of 10.7% in mAP@.50 and 13.0% in mAP@.50:.95. Compared to Cascade R-CNN, it was 31.0% higher in mAP@.50 and 18.9% higher in mAP@.50:.95. Against DDQ-4scale, the improvements were 10.3% for mAP@.50 and 12.2% for mAP@.50:.95. When compared to RTMDet-m, REAS-Det achieved a 9.4% increase in mAP@.50 and a 10.6% increase in mAP@.50:.95. The improvements over YOLOv8s were 4.1% in mAP@.50 and 5.5% in mAP@.50:.95, and compared to YOLOv8m, the gains were 2.9% for mAP@.50 and 3.8% for mAP@.50:.95. Lastly, against RT-DETR, the increases were 5.6% in mAP@.50 and 7.7% in mAP@.50:.95. Experimental results demonstrate that REAS-Det outperforms other advanced detection models in accuracy, particularly for Shatian pomelos detection in orchard environments, while maintaining a lightweight model size with only 5.8M parameters.

**Table 3. The result of comparative experiments**

| **Model** | **mAP@.50** | **mAP@.50:95** | **Parameters(M)** |
|---|---|---|---|
| Faster R-CNN | 53.7 | 20.2 | 40.3 |
| Rtmdet | 73.6 | 40.6 | 36.1 |
| CascadeR-CNN | 53.3 | 34.7 | 88.0 |
| DDQ-4scale | 74.0 | 41.4 | 42.0 |
| YOLOv8s | 80.2 | 48.1 | 11.2 |
| YOLOv8m | 81.4 | 49.8 | 25.9 |
| RTMDet-m | 74.9 | 43.0 | 24.7 |
| RT-DETR | 78.7 | 45.9 | 42.8 |
| REAS-Det | **84.3 (↑ 2.9%)** | **53.6 (↑ 3.8%)** | **5.8 (↓ 5.4%)** |

## 3.3 Discussion

To overcome the limitations of existing methods in real-world applications, four key variables that affect detection accuracy in practical applications were identified. A multi-strategy fruit detection framework was developed to address these challenges comprehensively. Experimental results confirmed that the proposed framework significantly improved the model's accuracy, while comparative analyses further demonstrated superior performance over other popular methods.

Notably, to closely simulate the real-world conditions of orchard object detection, many distant Shatian pomelos were inevitably captured in the images during data collection. These fruits were small and numerous, making it difficult for even humans to determine whether they were Shatian pomelos or not. Consequently, this affected the overall evaluation metrics, resulting in relatively lower mAP@0.50 and mAP@0.50:0.95 scores. However, in practice, Shatian pomelos far from the camera are not harvested immediately, so missed or incorrect detections of distant fruits have minimal impact on actual production environments. Additionally, the dataset includes a considerable number of low-resolution images, for which the model demonstrates suboptimal detection performance. This limitation highlights an important direction for future research in improving model robustness under low-quality visual conditions.

Lastly, the robustness and generalizability of the proposed algorithm still need further validation. The current study was limited to a single pomelo variety due to dataset availability constraints. Extending evaluation to diverse pomelo varieties remains challenging, primarily because of insufficient image data for many cultivars. Although this limitation restricts large-scale cross-variety testing, the algorithm's design suggests strong potential for broader applicability, particularly given the rapid advances in deep learning technologies.

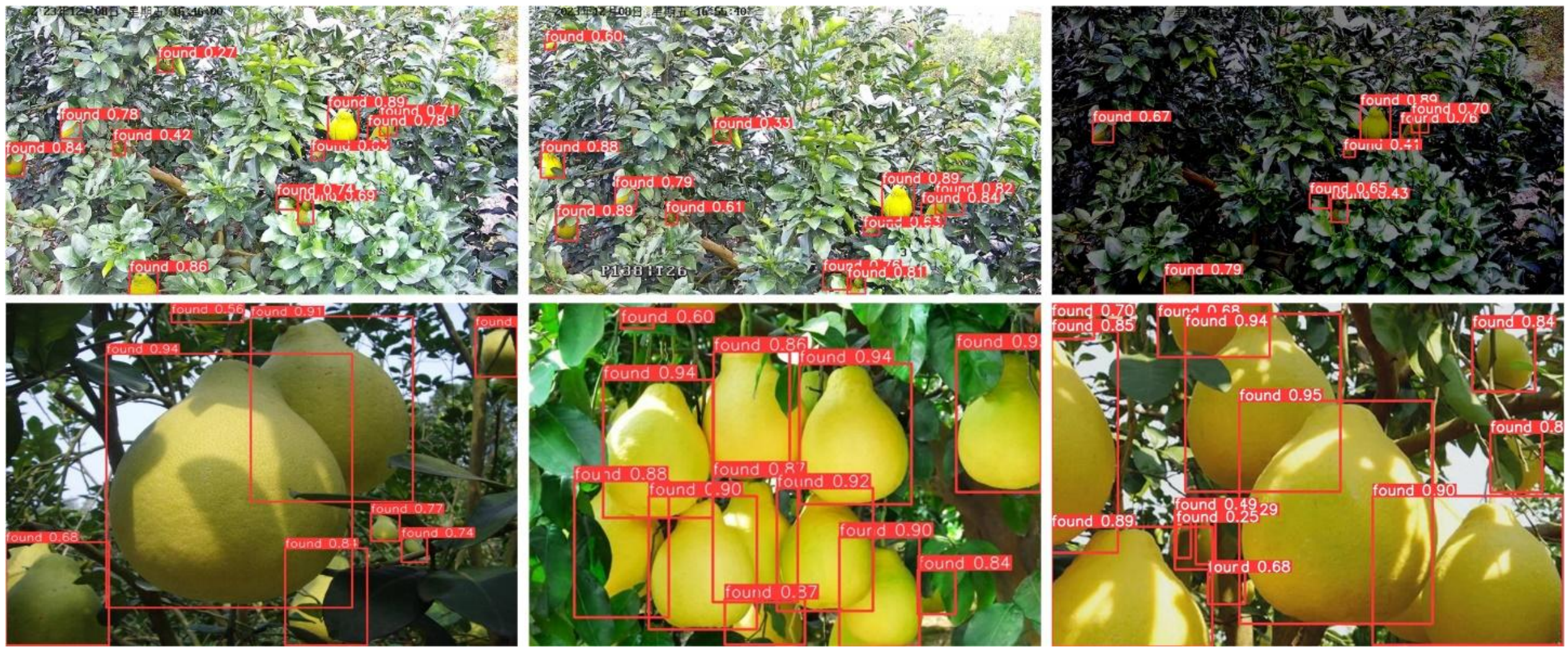

Fig. 11 illustrates the real-world detection results of the proposed framework under challenging conditions such as occlusion, low light, and scale variations.

## 4. Conclusion

Current fruit detection methodologies, while technically advanced, remain largely constrained to specific theoretical scenarios, which poses challenges when adapting these methods to real-world applications. Shatian pomelo, a fruit with regional significance and substantial market presence, faces four critical detection challenges in practical orchard environments. Overcoming these limitations through automated detection can yield considerable economic benefits and operational efficiencies. To accurately identify Shatian pomelos in real environments, a multi-strategy detection framework is developed to maintain consistent detection performance between field conditions and controlled images. The framework is designed as a multi-stage pipeline, with each stage strategically constructed to replicate real-world detection scenarios as closely as possible. It begins with specialized dataset construction, followed by targeted image augmentation, and culminates in an enhanced network architecture for robust detection performance. This network integrates the proposed GSV-Conv module, the C3RFEM module, the MultiSEAM module, and an improved NMS algorithm. The proposed GSV-Conv module represents a significant departure from traditional convolutional approaches by redefining feature visibility prior to aggregation. Unlike conventional methods that rely solely on receptive field expansion or post-aggregation attention mechanisms, GSV-Conv first expands the visible feature domain and then performs fine-grained spatial selection within the reconstructed visibility space. The MultiSEAM and C3RFEM modules play a crucial role in recognizing occluded Shatian pomelos, while replacing traditional NMS with soft-NMS further optimizes bounding box selection by imposing smoother penalties on overlapping regions. In addition, the proposed framework is evaluated in real orchard environments, where it demonstrates strong detection capability under natural conditions. Fig. 11 illustrates representative detection results, showing accurate identification of Shatian pomelos under challenging conditions such as occlusion, scale variation, and complex backgrounds.

In summary, the proposed framework introduces a novel and comprehensive approach to Shatian pomelo detection, with GSV-Conv providing a cutting-edge solution for efficient automated fruit counting in real-world orchard scenarios.

## CRediT authorship contribution statement

**Xiaodong Bai:** Conceptualization, Methodology, Formal analysis, Investigation, Writing – review & editing, Project administration, Funding acquisition. **Pan Wang:** Methodology, Writing – original draft, Writing – review & editing. **Yihao Hu:** Methodology, Validation, Data curation. **Jingchu Yang:** Writing – review & editing. **Leyi Zhou:** Writing – review & editing. **Aiping Yang:** Conceptualization, Formal analysis, Investigation, Resources. **Jianguo Yao:** Conceptualization, Project administration, Funding acquisition. **Xiangxiang Li:** Resources. **Meiping Ding:** Resources. **All authors:** Visualization, Editing.

## Declaration of Competing Interest

The authors declare that they have no known competing financial interests or personal relationships that could have appeared to influence the work reported in this paper.

## Acknowledgments

This work was supported in part by the National Natural Science Foundation of China (62462027 and 62271266), the Hainan Provincial Natural Science Foundation of China (625MS045), and the College Student Innovation Training Program Project (202410589043 and 202410589080).

## Data availability

Data will be made publicly available after the paper is published.

## References

Bodla, N., Singh, B., Chellappa, R., & Davis, L.S., 2017. Soft-NMS--improving object detection with one line of code, Proceedings of the IEEE international conference on computer vision. Publishing, pp. 5561-5569.

Cai, Z., & Vasconcelos, N. 2019. Cascade R-CNN: High quality object detection and instance segmentation. *IEEE transactions on pattern analysis and machine intelligence, 43*, 1483-1498. https://doi.org/10.1109/TPAMI.2019.2956516.

Chen, L., Gu, L., Li, L., Yan, C., & Fu, Y., 2025. Frequency Dynamic Convolution for Dense Image Prediction, Proceedings of the Computer Vision and Pattern Recognition Conference. Publishing, pp. 30178-30188.

Chen, W., Liu, M., Zhao, C., Li, X., & Wang, Y. 2024. MTD-YOLO: Multi-task deep convolutional neural network for cherry tomato fruit bunch maturity detection. *Computers and Electronics in Agriculture, 216*, 108533. https://doi.org/10.1016/j.compag.2023.108533.

Gai, R., Chen, N., & Yuan, H. 2023. A detection algorithm for cherry fruits based on the improved YOLO-v4 model. *Neural Computing and Applications, 35*, 13895-13906. https://doi.org/10.1007/s00521-021-06029-z.

Ganatra, N., & Patel, A. 2021. Deep learning methods and applications for precision agriculture. *Machine Learning for Predictive Analysis: Proceedings of ICTIS 2020*, 515-527. https://doi.org/10.1007/978-981-15-7106-0_51.

Jia, W., Wang, Z., Zhang, Z., Yang, X., Hou, S., & Zheng, Y. 2022. A fast and efficient green apple object detection model based on Foveabox. *Journal of King Saud University-Computer and Information Sciences, 34*, 5156-5169. https://doi.org/10.1016/j.jksuci.2022.01.005.

Jocher, G., Chaurasia, A., Stoken, A., Borovec, J., Kwon, Y., Michael, K., Fang, J., Yifu, Z., Wong, C., & Montes, D. 2022. ultralytics/yolov5: v7. 0-yolov5 sota realtime instance segmentation. *Zenodo*. https://doi.org/10.5281/zenodo.3908559.

Juneja, A., Juneja, S., Soneja, A., & Jain, S. 2021. Real time object detection using CNN based single shot detector model. *Journal of Information Technology Management, 13*, 62-80. https://doi.org/10.22059/jitm.2021.80025.

Kamilaris, A., & Prenafeta-Boldú, F.X. 2018. Deep learning in agriculture: A survey. *Computers and electronics in agriculture, 147*, 70-90. https://doi.org/10.1016/j.compag.2018.02.016.

Li, C., Zhou, A., & Yao, A. 2022. Omni-dimensional dynamic convolution. *arXiv preprint arXiv:2209.07947*.

Li, J., Wen, Y., & He, L., 2023. Scconv: Spatial and channel reconstruction convolution for feature redundancy,

Proceedings of the IEEE/CVF conference on computer vision and pattern recognition. Publishing, pp. 6153-6162.
Liu, J., & Wang, X. 2021. Plant diseases and pests detection based on deep learning: a review. *Plant Methods, 17*, 1-18. https://doi.org/10.1186/s13007-021-00722-9.
Liu, M., Jia, W., Wang, Z., Niu, Y., Yang, X., & Ruan, C. 2022. An accurate detection and segmentation model of obscured green fruits. *Computers and Electronics in Agriculture, 197*, 106984. https://doi.org/10.1016/j.compag.2022.106984.
Liu, S., Zeng, X., & Whitty, M. 2020. A vision-based robust grape berry counting algorithm for fast calibration-free bunch weight estimation in the field. *Computers and Electronics in Agriculture, 173*, 105360. https://doi.org/10.1016/j.compag.2020.105360.
Liu, T.-H., Ehsani, R., Toudeshki, A., Zou, X.-J., & Wang, H.-J. 2018. Identifying immature and mature pomelo fruits in trees by elliptical model fitting in the Cr–Cb color space. *Precision Agriculture, 20*, 138-156. https://doi.org/10.1007/s11119-018-9586-1.
Lyu, C., Zhang, W., Huang, H., Zhou, Y., Wang, Y., Liu, Y., Zhang, S., & Chen, K. 2022. Rtmdet: An empirical study of designing real-time object detectors. *arXiv preprint arXiv:2212.07784*.
Lyu, S., Li, R., Zhao, Y., Li, Z., Fan, R., & Liu, S. 2022. Green citrus detection and counting in orchards based on YOLOv5-CS and AI edge system. *Sensors, 22*, 576. https://doi.org/10.3390/s22020576.
Moussaid, A., El Fkihi, S., Zennayi, Y., Kassou, I., Bourzeix, F., Lahlou, O., El Mansouri, L., & Imani, Y. 2023. Citrus yield prediction using deep learning techniques: A combination of field and satellite data. *Journal of Open Innovation: Technology, Market, and Complexity, 9*, 100075. https://doi.org/10.1016/j.joitmc.2023.100075.
Newswire, P., 2023. Home of Shatian Pomelos in China: Rong County of Guangxi Promotes Improvement and Upgrade in Shatian Pomelo Industry. Publishing.
Oppenheim, D., Shani, G., Erlich, O., & Tsror, L. 2019. Using deep learning for image-based potato tuber disease detection. *Phytopathology, 109*, 1083-1087. https://doi.org/10.1094/PHYTO-08-18-0288-R.
Qi, J., Liu, X., Liu, K., Xu, F., Guo, H., Tian, X., Li, M., Bao, Z., & Li, Y. 2022. An improved YOLOv5 model based on visual attention mechanism: Application to recognition of tomato virus disease. *Computers and electronics in agriculture, 194*, 106780. https://doi.org/10.1016/j.compag.2022.106780.
Qi, Y., He, Y., Qi, X., Zhang, Y., & Yang, G., 2023. Dynamic snake convolution based on topological geometric constraints for tubular structure segmentation, Proceedings of the IEEE/CVF international conference on computer vision. Publishing, pp. 6070-6079.
Qiang, L., Jianrong, C., Bin, L., Lie, D., & Yajing, Z. 2014. Identification of fruit and branch in natural scenes for citrus harvesting robot using machine vision and support vector machine. *International Journal of Agricultural and Biological Engineering, 7*, 115-121. https://doi.org/10.3965/j.ijabe.20140702.014.
Ren, S., He, K., Girshick, R., & Sun, J. 2015. Faster r-cnn: Towards real-time object detection with region proposal networks. *Advances in neural information processing systems, 28*.
Sun, H., Li, Y., Li, Z., Yang, R., Xu, Z., Dou, J., Qi, H., & Chen, H. 2025. Fourier Convolution Block with global receptive field for MRI reconstruction. *Medical Image Analysis, 99*, 103349.
Sun, M., Xu, L., Luo, R., Lu, Y., & Jia, W. 2022. Fast location and recognition of green apple based on RGB-D image. *Frontiers in Plant Science, 13*, 864458. https://doi.org/10.3389/fpls.2022.864458.
Sunkara, R., & Luo, T., 2022. No more strided convolutions or pooling: A new CNN building block for low-resolution images and small objects, Joint European conference on machine learning and knowledge discovery in databases. Publishing, pp. 443-459.
Tian, Y., Yang, G., Wang, Z., Wang, H., Li, E., & Liang, Z. 2019. Apple detection during different growth stages in orchards using the improved YOLO-V3 model. *Computers and electronics in agriculture, 157*, 417-426.

https://doi.org/10.1016/j.compag.2019.01.012.

Wang, C.-Y., Yeh, I.-H., & Liao, H.-Y.M. 2024. Yolov9: Learning what you want to learn using programmable gradient information. *arXiv preprint arXiv:2402.13616*. https://doi.org/10.1007/978-3-031-72751-1_1.

Wang, J., Su, Y., Yao, J., Liu, M., Du, Y., Wu, X., Huang, L., & Zhao, M. 2023. Apple rapid recognition and processing method based on an improved version of YOLOv5. *Ecological Informatics, 77*, 102196. https://doi.org/10.1016/j.ecoinf.2023.102196.

Wang, X., Zheng, Z., Shao, J., Duan, Y., & Deng, L.-J., 2025. Adaptive Rectangular Convolution for Remote Sensing Pansharpening, Proceedings of the Computer Vision and Pattern Recognition Conference. Publishing, pp. 17872-17881.

Wen, C., Zhang, H., Li, H., Li, H., Chen, J., Guo, H., & Cheng, S., 2020. Multi-scene citrus detection based on multi-task deep learning network, 2020 IEEE International Conference on Systems, Man, and Cybernetics (SMC). Publishing, pp. 912-919. https://doi.org/10.1109/SMC42975.2020.9282909.

Xu, L., Zhu, S., Chen, X., Wang, Y., Kang, Z., Huang, P., & Peng, Y. 2020. Citrus recognition in real scenarios based on machine vision. *DYNA-Ingeniería e Industria, 95*. https://doi.org/10.6036/9363.

Yan, L., Jie, S., Hang, X., Guangyin, G., Jianxiong, L., & Jie, L. 2021. Detection and Grading Method of Pomelo Shape Based on Contour Coordinate Transformation and Fitting. *Smart Agriculture, 3*, 86. https://doi.org/10.12133/j.smartag.2021.3.1.202102-SA007.

Yang, R., Liu, J., Cao, S., Sun, W., & Kong, F. 2024. Impacts of agri-food E-commerce on traditional wholesale industry: Evidence from China. *Journal of Integrative Agriculture, 23*, 1409-1428. https://doi.org/10.1016/j.jia.2023.10.020.

Yu, Z., Huang, H., Chen, W., Su, Y., Liu, Y., & Wang, X. 2024. Yolo-facev2: A scale and occlusion aware face detector. *Pattern Recognition, 155*, 110714. https://doi.org/10.1016/j.patcog.2024.110714.

Zeng, Y.-F., Chen, C.-T., & Lin, G.-F. 2023. Practical application of an intelligent irrigation system to rice paddies in Taiwan. *Agricultural Water Management, 280*, 108216. https://doi.org/10.1016/j.agwat.2023.108216.

Zhang, S., Wang, X., Wang, J., Pang, J., Lyu, C., Zhang, W., Luo, P., & Chen, K., 2023. Dense distinct query for end-to-end object detection, Proceedings of the IEEE/CVF conference on computer vision and pattern recognition. Publishing, pp. 7329-7338.

Zhang, Y., Li, L., Chun, C., Wen, Y., Li, C., & Xu, G. 2024. Data-driven Bayesian Gaussian mixture optimized anchor box model for accurate and efficient detection of green citrus. *Computers and Electronics in Agriculture, 225*, 109366. https://doi.org/10.1016/j.compag.2024.109366.

Zhou, X., Zhang, Y., Jiang, X., Riaz, K., Rosenbaum, P., Lefsrud, M., & Sun, S. 2024. Advancing tracking-by-detection with MultiMap: Towards occlusion-resilient online multiclass strawberry counting. *Expert Systems with Applications, 255*, 124587. https://doi.org/10.1016/j.eswa.2024.124587.